\documentclass[conference]{IEEEtran}
\usepackage{cite}
\usepackage{amsmath,amssymb,amsfonts}
\usepackage{algorithmic}
\usepackage{graphicx}
\usepackage{textcomp}
\usepackage{xcolor}
\usepackage{orcidlink}

 \usepackage{booktabs}
 \usepackage{xurl}
\usepackage{hyperref}
 
\def\BibTeX{{\rm B\kern-.05em{\sc i\kern-.025em b}\kern-.08em
    T\kern-.1667em\lower.7ex\hbox{E}\kern-.125emX}}
\begin{document}

\title{LLM-Conditioned Synthesis of Pathological Gaits via Structured Gait-Language Representations\\
}

\author{
\IEEEauthorblockN{
Mritula Chandrasekaran\textsuperscript{1}\,\orcidlink{0000-0003-0866-6694},
Sanket Kachole\textsuperscript{2}\,\orcidlink{0000-0002-1496-2070},
Jarek Francik\textsuperscript{3}\,\orcidlink{0009-0003-8927-2802},
Dimitrios Makris\textsuperscript{4}\,\orcidlink{0000-0001-6170-0236}
}
\IEEEauthorblockA{
\textit{School of Computer Science and Mathematics},
\textit{Kingston University},
London, United Kingdom
}
\IEEEauthorblockA{
\textsuperscript{1,3,4}\{C.Mritula, jarek, D.Makris\}@kingston.ac.uk,
\textsuperscript{2}sanketkachole0707@gmail.com
}
}


\maketitle

\begin{abstract}
Pathological gait datasets remain scarce due to privacy, recruitment, cost, and movement variability. Our work presents a multimodal LLM-guided  framework for pathology-aware 3D gait data synthesis from structured textual descriptions. The proposed method generates fixed-length synthetic skeleton-based gait sequences for pathological gait classification tasks. The framework combines motion tokenisation, pathology-aware language conditioning, LLM-based semantic augmentation, and language-to-gait generation. A key contribution is the proposed pathological tokeniser, which is designed to preserve pathology-specific motion characteristics during discrete representation learning. Experiments suggest that the proposed synthetic sequences improve downstream classification for recurrent classifiers when combined with real data. The best result is obtained using a GRU classifier trained with real and synthetic samples, achieving 92.77\% accuracy under a leave-one-subject-out protocol.

\end{abstract}
Pathological gait analysis is important for clinical assessment, rehabilitation, and monitoring of movement disorders.  
However, labelled pathological gait data are limited because clinical data collection is costly, time-consuming, and subject to privacy constraints. Although generic text-to-motion and gait-synthesis methods can generate plausible motion, they do not preserve pathology-specific biomechanical cues \cite{tevet2022human, jiang2023motiongpt}. In pathological gait, relevant differences often appear through subtle changes in range of motion, stance, stride, asymmetry, and class-specific movement patterns \cite{jun2020pathological}. Existing motion-language and skeleton-augmentation methods may therefore increase sample diversity \cite{cormier2024enhancing} without preserving pathological signatures. Our work addresses this gap by introducing a pathology-aware gait-language synthesis framework with a dedicated pathological tokeniser. Beyond generating realistic gait sequences, our work aims to preserve discriminative pathological characteristics for downstream classification.



\begin{figure*}[t!]
  \centering  \includegraphics[width=0.7 \textwidth]{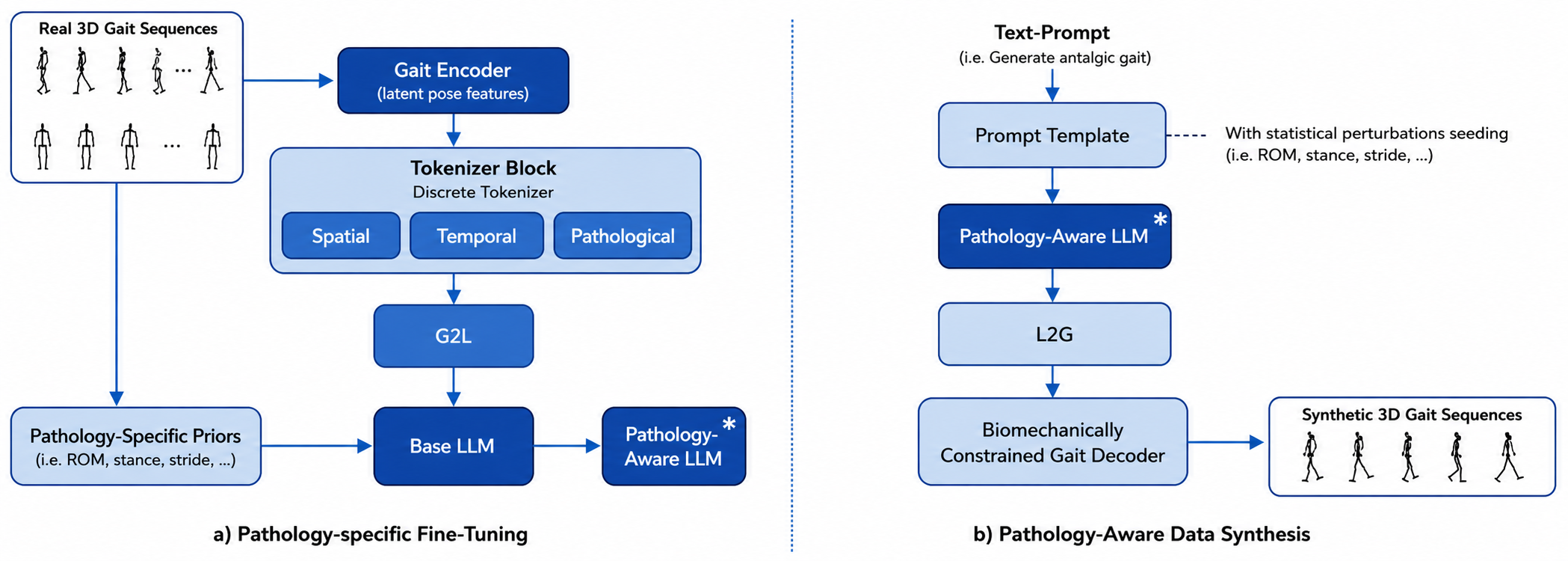}
  \caption{Proposed pathology-aware LLM based gait synthesis. (a) Real 3D gait sequences are encoded and discretised using spatial, temporal, and pathological tokenisers. Pathology-specific statistical priors, including ROM, stance, stride, asymmetry, and severity, are extracted from real data and used to guide LLM fine-tuning through G2L representations inspired by \cite{Yang2025GaitLLM}. (b) A text prompt is expanded through a prompt template using the learned pathology-specific priors. The pathology-aware LLM generates gait-language tokens, which are converted by the L2G module adapted from \cite{Yang2025GaitLLM} and decoded by a biomechanics-constrained gait decoder into synthetic 3D gait sequences.
}
  \label{fig:Pathology-awaregaitsynthesis}
\end{figure*}





\section{Methodology}
The proposed methodology is inspired by GaitLLM \cite{Yang2025GaitLLM}, which introduces Gait-to-Language and Language-to-Gait mappings for LLM-based gait sequence modelling. Unlike GaitLLM, which focuses on identity-based gait recognition using spatial and temporal tokens, the proposed framework targets pathology-aware 3D gait synthesis. It extends the tokenization stage through a dedicated pathological branch and conditions generation using gait-class-specific statistical priors. 

The proposed framework synthesises pathology-aware 3D skeletal gait sequences from structured textual descriptions. Given an input gait sequence, $\mathbf{X} \in \mathbb{R}^{T_{in} \times J \times 3}$, the objective is to generate a synthetic sequence $\hat{\mathbf{X}}$ that preserves pathology-discriminative motion characteristics while introducing controlled variation. The framework consists of pose encoding, pathology-aware tokenisation, gait-to-language (G2L) mapping, LLM fine-tuning, pathology-conditioned semantic generation, language-to-gait (L2G) reconstruction, and 3D gait decoding.

\label{eq:overall_synthesis}
    
First, each real gait sequence is passed through a PoseEncoder, which transforms 
3D joint coordinates into a compact latent representation. The encoder 
captures both spatial skeletal configuration and short-term gait dynamics, producing an encoded representation $\mathbf{Z}=E_p(\mathbf{X})$, where $E_p$ denotes the PoseEncoder.

The tokenisation stage consists of three complementary branches: spatial, temporal, and pathological tokenisation. The spatial branch captures skeletal configuration and inter-joint coordination, while the temporal branch models gait rhythm and motion continuity across consecutive frames. The proposed pathological branch is designed to preserve pathology-discriminative biomechanical characteristics, including restricted range of motion, stance variation, stride irregularity, asymmetry, and compensatory movements. The three token streams are fused to form the unified gait-token representation $\mathbf{Z}_{tok}$.

The fused gait tokens are mapped into a 
language-compatible representation $\mathbf{L}$ through the gait-to-language (G2L) module \cite{Yang2025GaitLLM}, producing a sequence of gait-language tokens that encode the gait class and associated motion descriptors.

For pathology-specific fine-tuning, the gait-language representation $\mathbf{L}$ is combined with statistical priors derived from real training data. 
The conditioning input to the base LLM is defined as:
\begin{equation}
\mathbf{C}_{p}=\Phi(\mathbf{L},\mathbf{P}_{p},y_{p}),
\end{equation}
where $\mathbf{P}_{p}$ denotes the pathology-specific priors, $y_{p}$ the pathology class label, and $\Phi(\cdot)$ formats these components into a prompt-compatible representation. The base LLM is fine-tuned using $\mathbf{C}_{p}$, producing a pathology-aware model $f_{\theta_{LLM}}$.

During synthesis, the target gait class is specified. Then the corresponding pathology-specific priors are retrieved from the training data and used to construct a structured synthesis prompt, $\mathcal{P}_{syn}$. This prompt is passed to the fine-tuned pathology-aware LLM, $f_{\theta_{LLM}}$, which generates a pathology-conditioned language-token sequence, $\widetilde{\mathbf{L}}_{tok}$. This stage is guided by dataset-derived priors, ensuring that the generated representation remains pathology-plausible and aligned with the selected pathological gait class.

The generated language-token sequence $\widetilde{\mathbf{L}}_{tok}$ is then passed to the language-to-gait (L2G) module adapted from \cite{Yang2025GaitLLM}, which maps it back into a synthetic motion-token representation, $\widetilde{\mathbf{Z}}_{tok}$. This stage adapts the language-to-gait direction for generative synthesis, producing pathology-conditioned motion tokens.
The synthetic motion tokens are then projected into a decoder-compatible latent space, $\widetilde{\mathbf{H}}$. Finally, the gait decoder reconstructs the continuous 3D skeletal gait sequence. 

\section{Experiments}

The experiments are conducted on the Pathological Gait Dataset by Jun et al.~\cite{jun2020pathological}. 
The evaluation focuses on pathological gait synthesis using GPT-2 following Yang. et.al\cite{Yang2025GaitLLM}  and classification using GRU \cite{yu2021improved}, LSTM \cite{lin2020framework}, and CNN \cite{nguyen2023human} classifiers. Classification accuracy is reported in Table \ref{tab:classification_pt_results}. 

\section{Results and Discussion}

\begin{table}[h!]
\centering
\resizebox{\linewidth}{!}{
\begin{tabular}{l c c c}
\toprule
\textbf{Classifier} & \textbf{Only Real} & \textbf{Synthetic Only (With PT)} & \textbf{Real + Synthetic (With PT)} \\
\midrule
GRU  & 91.08 & 85.63 & \textbf{\underline{92.77}} \\
LSTM & 88.67 & 83.16 & \textbf{89.23} \\
CNN  & 90.17 & 79.00 & 87.97 \\
\bottomrule \newline
\end{tabular}
}

\caption{Classification accuracy (\%)  using real data and synthetic data generated with the Pathological Tokeniser (PT). }
\label{tab:classification_pt_results}
\end{table}

The GRU classifier performance improves from 91.08\% with real data only to 92.77\% with real and synthetic data. The LSTM shows a smaller positive gain, whereas the CNN decreases from 90.17\% to 87.97\% after adding synthetic data. This indicates that the recurrent models are better suited to learning from temporally consistent synthetic gait patterns. 

\begin{table}[h!]
\centering
\resizebox{\linewidth}{!}{
\begin{tabular}{l c c c}
\toprule
\textbf{LLM Model} 
& \textbf{Qwen-5B}
& \textbf{MotionGPT Synthetic} 
& \textbf{Proposed Method} \\
\midrule
\textbf{Accuracy (\%)} & 79.86 & 90.26 & \textbf{\underline{92.77}} \\
\bottomrule \newline
\end{tabular}
}
\caption{Comparative results. using a GRU gait classifier trained with data synthesised by different LLMs. }
\label{tab:motiongpt_comparison}
\end{table}


Table~\ref{tab:motiongpt_comparison} reports a comparative baseline evaluation against MotionGPT~\cite{ban2026diffusion,ribeiro2024motiongpt} and Qwen-5B~\cite{bai2023qwen}. These baselines are included to provide an initial comparison with existing general-purpose motion-language and LLM-based generation frameworks, as, to the best of our knowledge, no dedicated pathology-aware gait synthesis model currently exists. The proposed fine-tuned GPT-2 framework achieves 92.77\% classification accuracy, compared with 90.26\% for MotionGPT-generated synthetic data and 79.86\% for Qwen-5B.

\section{Conclusion}
\label{Conclusion}

This work introduced a multimodal LLM-guided framework for pathology-aware 3D gait synthesis using structured gait-language representations. 
The method combines motion tokenisation, pathology-aware prompting, semantic augmentation, and language-to-gait generation with a dedicated pathological tokeniser. 
Experimental results show that synthetic data can improve sequence-based classification when combined with real data, with the best GRU result reaching 92.77\% under LOSO evaluation. 
Comparisons with Qwen-5B and MotionGPT provide initial evidence for the benefit of pathology-aware conditioning over general-purpose generation baselines. 
Future work will focus on statistical validation, biomechanical assessment, expert review, and engaging  larger clinical datasets.

\vspace{12pt}
\color{red}


\begin{thebibliography}{00}

\bibitem{ribeiro2024motiongpt}
J. Ribeiro-Gomes, T. Cai, Z. A. Milacski, C. Wu, A. Prakash, S. Takagi, A. Aubel, D. Kim, A. Bernardino, and F. De La Torre, ``MotionGPT: Human motion synthesis with improved diversity and realism via GPT-3 prompting,'' in \textit{Proc. IEEE/CVF Winter Conf. Appl. Comput. Vis. (WACV)}, 2024, pp. 5058--5068, doi: 10.1109/WACV57701.2024.00499.


\bibitem{Yang2025GaitLLM}
W. Yang, S. Wang, J. Hou, H. Liu, C. Cao, and K. Huang, 
``Bridging gait recognition and large language models sequence modeling,'' 
in \textit{Proc. IEEE/CVF Conf. Comput. Vis. Pattern Recognit. (CVPR)}, 2025. 
[Online]. Available: \url{https://openaccess.thecvf.com/content/CVPR2025/html/Yang_Bridging_Gait_Recognition_and_Large_Language_Models_Sequence_Modeling_CVPR_2025_paper.html}

\bibitem{jun2020pathological}
K. Jun, Y. Lee, S. Lee, D.-W. Lee, and M. S. Kim, 
``Pathological gait classification using Kinect v2 and gated recurrent neural networks,'' 
\textit{IEEE Access}, vol. 8, pp. 139881--139891, 2020.

\bibitem{lin2020framework}
C.-B. Lin, Z. Dong, W.-K. Kuan, and Y.-F. Huang, 
``A framework for fall detection based on OpenPose skeleton and LSTM/GRU models,'' 
\emph{Applied Sciences}, vol. 11, no. 1, p. 329, 2020.

\bibitem{nguyen2023human}
K. Nguyen, V. V. Nguyen, N. T. Mai, A. H. Nguyen, and A. V. Nguyen,
``Human gait analysis using hybrid convolutional neural networks,''
\emph{Journal of Computer Science and Cybernetics}, vol. 39, no. 2, pp. 125--142, 2023.

\bibitem{bai2023qwen}
J. Bai, S. Bai, Y. Chu, Z. Cui, K. Dang, X. Deng, Y. Fan, W. Ge, Y. Han, F. Huang, \emph{et al.},
``Qwen technical report,''
\emph{arXiv preprint arXiv:2309.16609}, 2023.

\bibitem{ban2026diffusion}
J. Ban, J. Jeon, and S.. Jeong,
``From diffusion to flow: Efficient motion generation in MotionGPT3,''
\emph{arXiv preprint arXiv:2603.26747}, 2026.

\bibitem{yu2021improved}
W. Yu, R. Liu, D. Zhou, Q. Zhang, and X. Wei,
``An improved GRU network for human motion prediction,''
in \emph{Proc. 2021 IEEE 7th Int. Conf. Virtual Reality (ICVR)}, 2021, pp. 427--433.


\bibitem{tevet2022human}
G. Tevet, S. Raab, B. Gordon, Y. Shafir, D. Cohen-Or, and A. H. Bermano,
``Human Motion Diffusion Model,''
\emph{arXiv preprint arXiv:2209.14916}, 2022.

\bibitem{jiang2023motiongpt}
B. Jiang, X. Chen, W. Liu, J. Yu, G. Yu, and T. Chen,
``MotionGPT: Human Motion as a Foreign Language,''
in \emph{Advances in Neural Information Processing Systems}, 2023.

\bibitem{cormier2024enhancing}
M. Cormier, H. F. G. Nunes, and J. Beyerer,
``Enhancing Skeleton-Based Action Recognition in Real-World Scenarios Through Realistic Data Augmentations,''
in \emph{Proc. IEEE/CVF Winter Conference on Applications of Computer Vision Workshops (WACVW)}, 2024.


\end{thebibliography}
\end{document}